\documentclass[sigconf]{acmart}
\AtBeginDocument{%
  \providecommand\BibTeX{{%
    \normalfont B\kern-0.5em{\scshape i\kern-0.25em b}\kern-0.8em\TeX}}}

\copyrightyear{2023}
\acmYear{2023}
\setcopyright{acmlicensed}\acmConference[BuildSys '23]{The 10th ACM International Conference on Systems for Energy-Efficient Buildings, Cities, and Transportation}{November 15--16, 2023}{Istanbul, Turkey}
\acmBooktitle{The 10th ACM International Conference on Systems for Energy-Efficient Buildings, Cities, and Transportation (BuildSys '23), November 15--16, 2023, Istanbul, Turkey}
\acmPrice{15.00}
\acmDOI{10.1145/3600100.3623730}
\acmISBN{979-8-4007-0230-3/23/11}

\usepackage{subfigure}
\usepackage{multirow}

\begin{document}

\title{Utilizing Language Models for Energy Load Forecasting}

\author{Hao Xue}
\email{hao.xue1@unsw.edu.au}
\affiliation{%
  \institution{University of New South Wales}
  \city{Sydney}
  \state{NSW}
  \country{Australia}
}
\author{Flora D. Salim}
\email{flora.salim@unsw.edu.au}
\affiliation{%
  \institution{University of New South Wales}
  \city{Sydney}
  \state{NSW}
  \country{Australia}
}

\renewcommand{\shortauthors}{Xue and Salim}

\newcommand{\ie}{\textit{i.e.}}
\newcommand{\eg}{\textit{e.g.}}
\begin{abstract}
  Energy load forecasting plays a crucial role in optimizing resource allocation and managing energy consumption in buildings and cities. In this paper, we propose a novel approach that leverages language models for energy load forecasting. We employ prompting techniques to convert energy consumption data into descriptive sentences, enabling fine-tuning of language models. By adopting an autoregressive generating approach, our proposed method enables predictions of various horizons of future energy load consumption. Through extensive experiments on real-world datasets, we demonstrate the effectiveness and accuracy of our proposed method. Our results indicate that utilizing language models for energy load forecasting holds promise for enhancing energy efficiency and facilitating intelligent decision-making in energy systems.
\end{abstract}




\maketitle

\section{Introduction}
\begin{figure}
    \centering
    \subfigure[Typical numerical-based forecasting]{\includegraphics[width=0.4\textwidth]{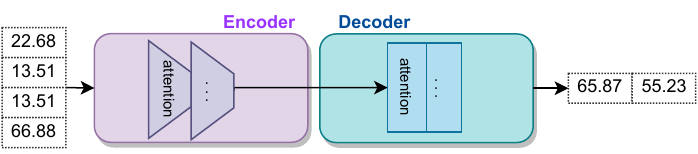}} \\ \vspace{-3ex}
  \subfigure[The proposed language-based forecasting]{\includegraphics[width=0.3\textwidth]{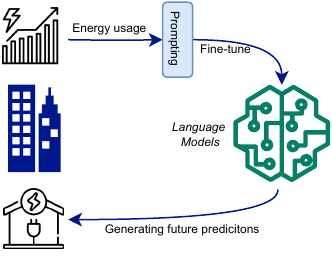}}
    \caption{The concept illustration of: (a) the  numerical-based deep learning forecasting methods; and (b) the proposed approach of using language models to forecast energy load.}
    \label{fig:intro}
\end{figure}

With the increasing need for energy efficiency and sustainable resource management, the forecasting of energy load has become a critical requirement in buildings, cities, and transportation systems. Accurate load forecasting enables proactive resource allocation, optimal demand response, and efficient energy management. 
Traditional approaches to energy load forecasting typically rely on statistical models and recent deep learning-based time series analysis techniques.
In recent years, language models based on deep learning, particularly Transformer-based models, have shown remarkable performance in various natural language processing tasks. These models have the ability to learn rich representations of textual data and capture intricate relationships between words and concepts. Typically, the forecasting process of these deep learning models involves one encoder that takes a sequence of numbers that stands for the historical energy consumption values as input and one decoder to generate another sequence of numerical values as the predicted future energy data, as illustrated in Figure~\ref{fig:intro} (a).

Motivated by the success of language models in natural language processing, we propose to leverage their power for energy load forecasting.
As demonstrated in Figure~\ref{fig:intro} (b), the core of our approach is converting energy consumption data into natural language sentences using prompting techniques. By describing the data as sentences, we aim to unlock the potential of language models to capture nuanced patterns and dependencies within the data. This representation allows us to fine-tune pre-trained language models, enabling them to learn from the specific characteristics of energy consumption sequences.
Similar numerical prompting has been used for human mobility data~\cite{xue2022translating,xue2022leveraging} recently. However, their prompts only support the forecasting of the next time step, which is limited for predicting the future energy load.
To this end, we further introduce an autoregressive mechanism in the prediction generation process with the fine-tuned language models.
This approach allows us to generate predictions for different horizons, ranging from short-term (\eg, the next time step) to long-term (\eg, the next 24 time steps) load forecasts.

Our method presents a novel ``code less'' solution for energy load forecasting, which could provide a new perspective rather than focusing on designing complicated deep learning forecasting models (\eg, Transformer-based methods).
This would makes it a relatively easy-accessible and user-friendly method for non-AI users, compared to existing forecasting models that require many tedious parameter searching and training processes.
In summary, our main contributions in this work are twofold:
    (1) We present a study on the utilization of language models for energy load forecasting. We design a pipeline that converts the energy consumption data into sentences for fine-tuning the language models and leverages the autoregressive mechanism for predicting different horizons with the same fine-tuned model.
    (2) We provide a comprehensive evaluation of the proposed solution with real world data from 6 buildings. We also conduct different evaluation settings including zero-shot performance evaluation and varied prediction horizon evaluation.

\section{Forecasting with Language Models}\label{sec:2}

\subsection{Problem Formulation and Method Overview}

Assume that the energy consumption records of a building $i$ is represented by a sequence of $t$ continuous time steps $\mathcal{X}^i= \{x_1^i ,x_2^i, \cdots, x_t^i\}$. The value indicates that the energy consumption of building $i$ at time $t$ is $x_t^i$.
The energy load forecasting problem can then be formulated as predicting the future load consumption values $y^i_{t_1: t_{m}}$ of the next $m$ time steps given the history observation $x^i_{t_1: t_{n}}$. Here, $n$ and $m$ are the observation length and the prediction horizon.

Overall, as illustrated in Figure~\ref{fig:intro} (b), the proposed method comprises three key enablers: (1) \textbf{Prompting}: to transform the raw consumption data into sentences that can be processed by language models; (2) \textbf{Fine-tuning language models}: to adapt them to the specifics of energy forecasting task; (3) \textbf{Autoregressive generation}: to enable the generation of multiple future steps forecasting.

\subsection{Prompting and Fine-tuning}


To utilize language models for energy load forecasting, we employ a prompting technique that translates the usage data.
Generally, the raw energy data is provided in a tabular format and we translate each row into a descriptive sentence. The objective is to transform the raw numerical data into a natural language text format that is suitable for language models and captures the relevant information and context necessary for predictions. By converting the energy consumption data into sentences, language models are enabled to take the transformed energy data as input and capture nuanced patterns and dependencies within the data.

In the prompting process, we utilize a predefined template (\eg, ``The electric load at \{Time\} is \{Usage\}.'') that serves as a backbone for constructing the sentences. The template consists of placeholders for the actual values from the data, resulting in sentences that convey the energy consumption information in a human-readable format. The template includes variables such as date, time, energy consumption, and any other metadata provided in the raw data that may be relevant for forecasting.
By replacing the placeholders in the template with the actual values, we obtain a sentence that represents the energy consumption data for a particular time step. This process is repeated for each row in the raw tabular data, resulting in a collection of descriptive sentences that are used for fine-tuning the language models.
Through prompting, we bridge the gap between numerical energy consumption data and the language model's ability to comprehend and generate textual information.

After generating the sentences from the energy consumption data, we proceed to fine-tune a pre-trained language model. Fine-tuning allows the model to adapt to the specific characteristics of load forecasting and capture the dependencies within the data.
The language models are often pre-trained on a large corpus of text data to learn general language representations and common knowledge.
In this study, we leverage the pre-trained models provided by 
\href{https://huggingface.co/models}{HuggingFace}
and the models are pre-trained only with general English-language corpora datasets without any specific energy usage-related numerical datasets.
Fine-tuning involves training the language model on our generated load sentences to specialize it for load forecasting. During the fine-tuning process, we feed the generated sentences as input to the language models and optimize the parameters to minimize the difference between the predicted next sentence and the ground truth sentence. This process could make the language models suitable for our load forecasting task.

\subsection{Autoregressive Generation}
Taking a set of historical sentences representing past energy load consumption as input, the fine-tuned language models generate the next sentence, which corresponds to the predicted energy load for the next time step.
When longer prediction horizons are required, we adopt an autoregressive method and we extend the prediction horizon by appending the generated sentence to the end of the input sentence sequence. This extended sequence becomes the input for predicting the subsequent time step. By iteratively generating sentences in an autoregressive manner, we can forecast energy load consumption for multiple future time steps.

The autoregressive generation approach leverages the fine-tuned language model's ability to capture dependencies between historical consumption and future load patterns. By using the generated sentences as input, the model can adjust its predictions based on the evolving context, enabling dynamic forecasts for different horizons.

\section{Experiments}

\subsection{Dataset and Evaluation}
This study uses data from a certain block within the Melbourne CBD area in Australia.
We select aggregated and anonymised smart meter data of hourly energy consumption for 6 buildings (\ie, Building A-F). The data is collected from 2018 January to 2019 December. For each building, the data of the first 22 months is considered as the training set to fine-tune the language models. The data of the last month (Dec. 2019) is used as the testing set and the remaining month (Nov. 2019) is split as the validation set.

\begin{table*}[]
\centering
\caption{The comparison between numerical forecasting methods and the proposed forecasting with language models.}
\label{tab:res}
\footnotesize
\begin{tabular}{|l|cc|cc|cc|cc|cc|cc|} \hline
 & \multicolumn{2}{c|}{Building A} & \multicolumn{2}{c|}{Building B} & \multicolumn{2}{c|}{Building C} & \multicolumn{2}{c|}{Building D} & \multicolumn{2}{c|}{Building E} & \multicolumn{2}{c|}{Building F} \\ \cline{2-13}
 & RMSE & MAE & RMSE & MAE & RMSE & MAE & RMSE & MAE & RMSE & MAE & RMSE & MAE \\ \hline \hline
Transformer & 143.383 & 110.462 & 6.022 & 3.966 & 47.671 & 30.051 & 18.008 & 12.932 & 6.950 & 4.735 & 37.636 & 26.628 \\
Informer & 176.335 & 135.877 & 5.127 & 3.274 & 45.660 & 28.253 & \textbf{17.557} & \textbf{12.408} & 5.999 & 4.194 & 35.066 & 24.261 \\
Autoformer & 60.703 & 41.873 & 5.521 & 3.891 & 54.415 & 38.463 & 26.603 & 18.902 & 8.335 & 6.142 & 45.291 & 34.578 \\
FEDformer & 50.055 & 32.488 & 5.455 & 3.772 & 48.001 & 33.619 & 27.548 & 20.789 & 7.882 & 5.879 & 44.008 & 33.883 \\ \hline
\multicolumn{13}{|c|}{Language Model-based (ours)} \\ \hline
Bart & 28.634 & 20.489 & 3.266 & 1.871 & 33.635 & 25.050 & 48.555 & 28.074 & \textbf{4.308} & \textbf{2.793} & 28.734 & \textbf{20.990} \\
Bigbird & 28.350 & 19.181 & 4.617 & 2.803 & \textbf{25.466} & \textbf{19.903} & 34.831 & 19.235 & 4.402 & 2.988 & 28.509 & 21.237 \\
Pegasus & \textbf{20.419} & \textbf{15.124} & \textbf{3.186} & \textbf{1.865} & 29.453 & 22.201 & 53.986 & 32.419 & 5.161 & 3.201 & \textbf{28.434} & 21.298 \\ \hline
\end{tabular}
\end{table*}

\begin{table*}[]
\centering
\caption{Results of language models under the zero-shot setting. }
\label{tab:zero}
\footnotesize
\begin{tabular}{|l|l|cc|cc|cc|cc|cc|} \hline
\multirow{2}{*}{Fine-tuning data} &  & \multicolumn{2}{c}{Building B} & \multicolumn{2}{c}{Building C} & \multicolumn{2}{c}{Building D} & \multicolumn{2}{c}{Building E} & \multicolumn{2}{c}{Building F} \\ \cline{3-12}
 &  & RMSE & MAE & RMSE & MAE & RMSE & MAE & RMSE & MAE & RMSE & MAE \\ \hline
 & Bart & 49.645 & 9.780 & 53.534 & 40.466 & \textbf{18.154} & \textbf{12.276} & 142.760 & 65.407 & 60.022 & 43.923 \\
Building A & Bigbird & 140.140 & 104.256 & \textbf{34.929} & \textbf{24.990} & \textbf{20.014} & \textbf{13.028} & 193.763 & 183.637 & 51.590 & 36.235 \\
 & Pegasus & 71.213 & 39.008 & \textbf{27.545} & \textbf{21.400} & \textbf{22.655} & \textbf{15.344} & 178.871 & 140.538 & 40.983 & 32.209 \\ \hline \hline
\multicolumn{2}{|l|}{} & \multicolumn{2}{c|}{Building A} & \multicolumn{2}{c|}{Building C} & \multicolumn{2}{c|}{Building D} & \multicolumn{2}{c|}{Building E} & \multicolumn{2}{c|}{Building F} \\ \hline
 & Bart & 265.186 & 258.663 & 223.216 & 215.495 & 121.805 & 116.927 & 10.050 & 7.318 & 275.279 & 262.617 \\
Building B & Bigbird & 245.885 & 239.373 & 207.996 & 199.670 & 115.281 & 104.371 & 6.404 & 4.189 & 255.267 & 241.459 \\ 
 & Pegasus & 209.142 & 180.299 & 182.130 & 158.734 & 107.290 & 95.584 & \textbf{5.568} & \textbf{4.053} & 222.276 & 192.440 \\ \hline \hline
\multicolumn{2}{|l|}{} & \multicolumn{2}{c|}{Building A} & \multicolumn{2}{c|}{Building B} & \multicolumn{2}{c|}{Building D} & \multicolumn{2}{c|}{Building E} & \multicolumn{2}{c|}{Building F} \\ \hline
 & Bart & \textbf{31.098} & \textbf{24.549} & 11.771 & 4.609 & 121.081 & 100.328 & 52.622 & 15.383 & 40.697 & 30.991 \\
Building C & Bigbird & \textbf{31.565} & \textbf{25.242} & 16.049 & 4.201 & 53.462 & 37.821 & 147.580 & 97.219 & 36.477 & 27.524 \\
 & Pegasus & \textbf{31.859} & \textbf{25.333 }& 11.736 & 6.661 & 79.377 & 64.420 & 10.332 & 5.932 & 35.590 & 27.711 \\ \hline \hline
\multicolumn{2}{|l|}{} & \multicolumn{2}{c|}{Building A} & \multicolumn{2}{c|}{Building B} & \multicolumn{2}{c|}{Building C} & \multicolumn{2}{c|}{Building E} & \multicolumn{2}{c|}{Building F} \\ \hline
 & Bart & 103.643 & 77.086 & 60.994 & 21.930 & 68.729 & 49.655 & 90.899 & 48.022 & 134.490 & 95.149 \\
Building D & Bigbird & 37.347 & 26.154 & 85.492 & 49.806 & \textbf{36.092} & \textbf{26.142} & 182.591 & 168.000 & 68.114 & 49.591 \\
 & Pegasus & 57.576 & 41.751 & 35.009 & 10.887 & 47.066 & 35.939 & 76.964 & 28.347 & 82.010 & 59.684 \\ \hline \hline
\multicolumn{2}{|l|}{} & \multicolumn{2}{c|}{Building A} & \multicolumn{2}{c|}{Building B} & \multicolumn{2}{c|}{Building C} & \multicolumn{2}{c|}{Building D} & \multicolumn{2}{c|}{Building F} \\ \hline
 & Bart & 246.991 & 239.025 & 9.345 & 7.148 & 204.619 & 195.582 & 97.547 & 89.958 & 258.007 & 243.845 \\
Building E & Bigbird & 251.074 & 245.359 & 34.759 & 27.898 & 211.649 & 203.910 & 82.280 & 63.093 & 260.483 & 247.872 \\
 & Pegasus & 240.030 & 228.999 & 9.827 & 4.740 & 192.677 & 170.909 & 70.708 & 51.297 & 251.381 & 232.470 \\ \hline \hline
\multicolumn{2}{|l|}{} & \multicolumn{2}{c|}{Building A} & \multicolumn{2}{c|}{Building B} & \multicolumn{2}{c|}{Building C} & \multicolumn{2}{c|}{Building D} & \multicolumn{2}{c|}{Building E} \\ \hline
 & Bart & 40.445 & 32.498 & 15.492 & 4.670 & \textbf{27.708} & \textbf{20.957} & 55.216 & 35.233 & 56.668 & 15.612 \\
Building F & Bigbird & 34.152 & 27.685 & 11.730 & 3.715 & \textbf{26.631} & \textbf{20.123} & \textbf{28.173} & \textbf{18.846} & 162.201 & 113.104 \\
 & Pegasus & 39.271 & 32.015 & 7.077 & 3.802 & \textbf{30.097} & \textbf{23.107} & 38.075 & 25.148 & 12.781 & 6.263 \\ \hline
\end{tabular}
\end{table*}

For evaluating the performance of each method, we use the Root Mean Squared Error (RMSE) and the Mean Absolute Error (MAE) as metrics. For both the measures, the lower error means the better performance.
These errors are calculated based on the predicted $\hat{y}_{t_{1}:t_{m}}^i$ and the ground truth $y_{t_{1}:t_{m}}^i$ to measure the closeness of the predicted values.
For this study, we have fixed a forecast horizon size of $m=24$ hours, to mimic a day-ahead forecasting experiment. To make the input observation size bigger than the output prediction horizon, we set the input observation length $n=30$. 


\subsection{Performance}

\subsubsection{Comparing Against Numerical Forecasting Methods}

To evaluate the performance of our approach, we compare it against the typical numerical forecasting methods commonly used for time series forecasting. Specifically, we select the popular Transformer~\cite{vaswani2017attention} as well as the more recent Informer~\cite{zhou2021informer}, Autoformer~\cite{xu2021autoformer}, and FEDformer~\cite{FEDformer} as numerical baselines.
For language models, a recent benchmark study~\cite{xue2023promptcast} has shown that three language models (Bart~\cite{bart},
Bigbird~\cite{zaheer2020big}, and Pegasus~\cite{zhang2020pegasus}) have better forecasting ability. These models also have reasonable size and number of parameters so that they can run on a single GPU (\eg, we used Nvidia V100 in our experiments).
Thus, these three language models are selected in our evaluation and our implementations are available at: \url{https://github.com/xuehaouwa/LM-Load-Forecasting}.

The best performance under each column is shown in bold in Table~\ref{tab:res}.
From the table, it is evident that language models, with our proposed pipeline, outperform traditional numerical forecasting methods in the majority of cases.
Specifically, the language models have superior performance over baselines with a significant gain on Building A, B, C, and F.
These findings highlight the language models' ability to capture patterns within energy consumption data, ultimately leading to more accurate predictions.

\begin{figure*}
  \centering
  
  \subfigure{\includegraphics[width=0.275\textwidth]{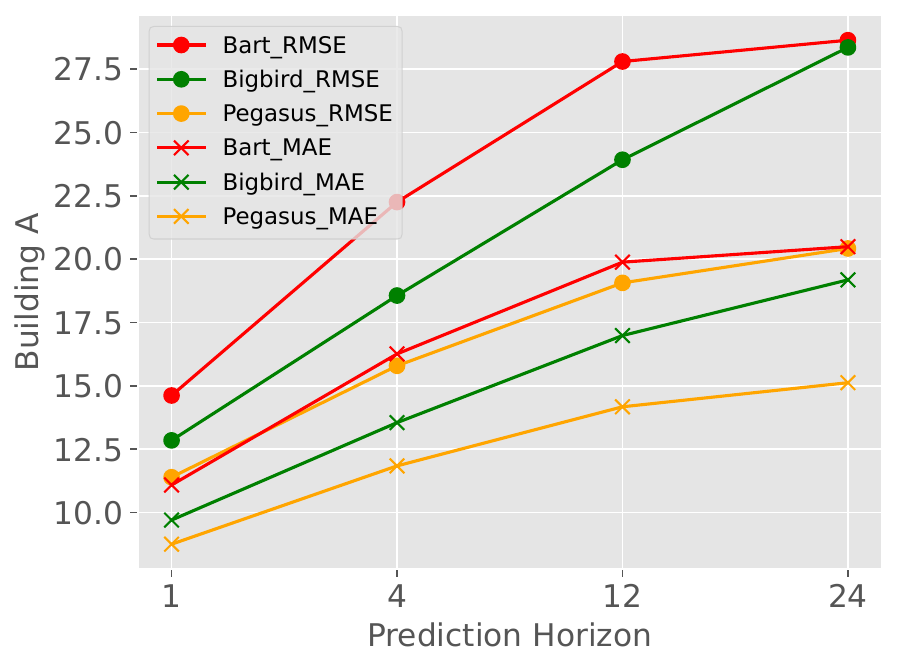}}
  \subfigure{\includegraphics[width=0.275\textwidth]{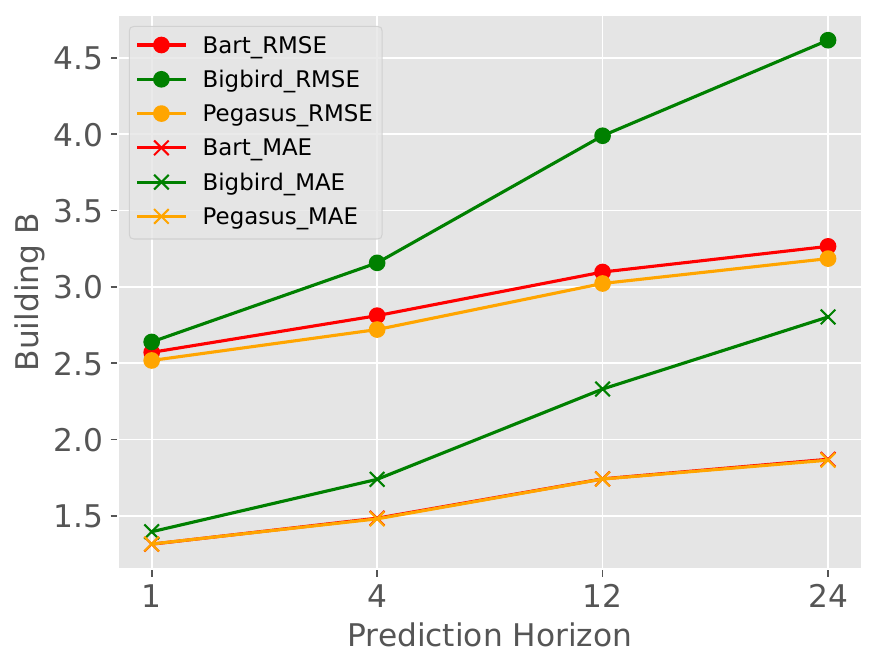}}
  \subfigure{\includegraphics[width=0.275\textwidth]{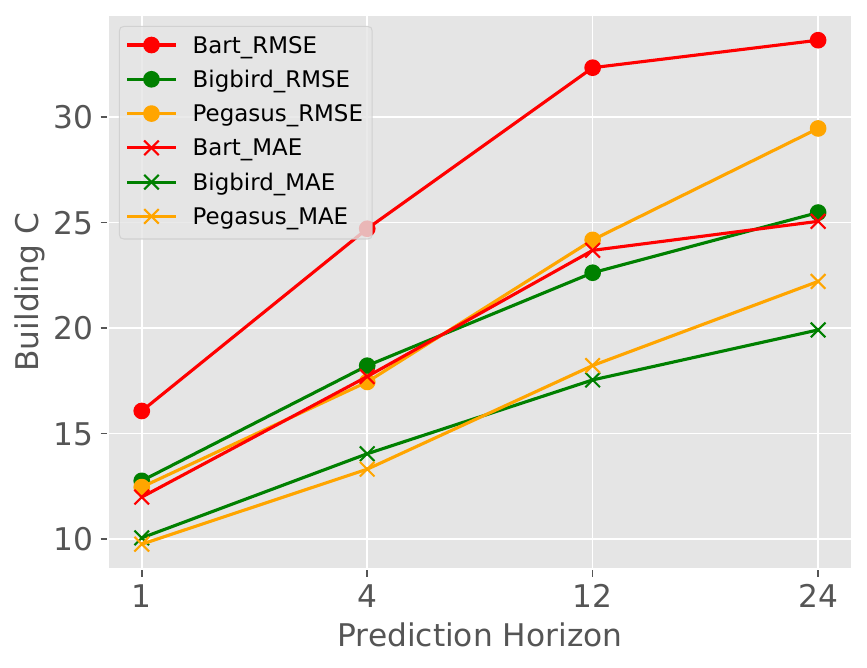}} \\ \vspace{-3ex}
  \subfigure{\includegraphics[width=0.275\textwidth]{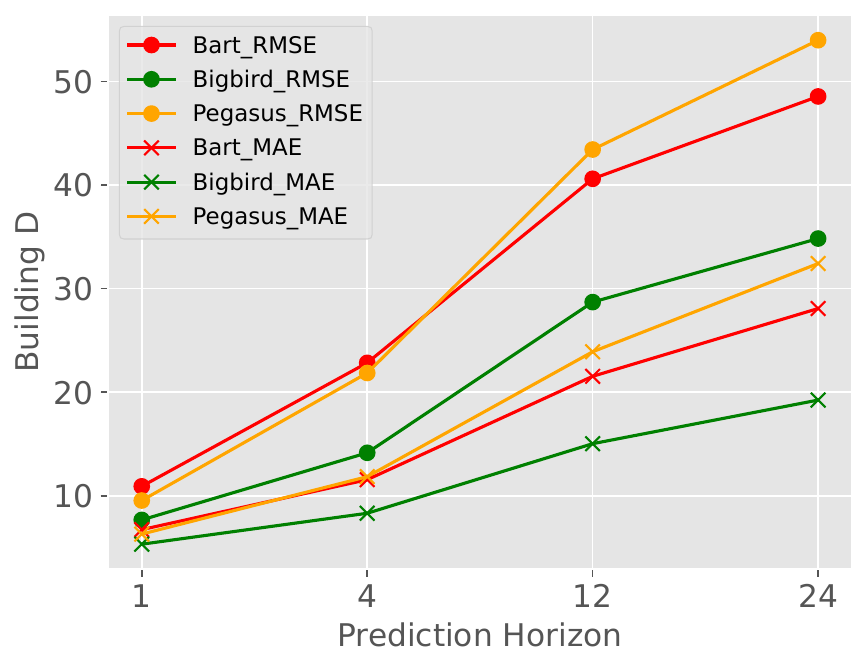}}
  \subfigure{\includegraphics[width=0.275\textwidth]{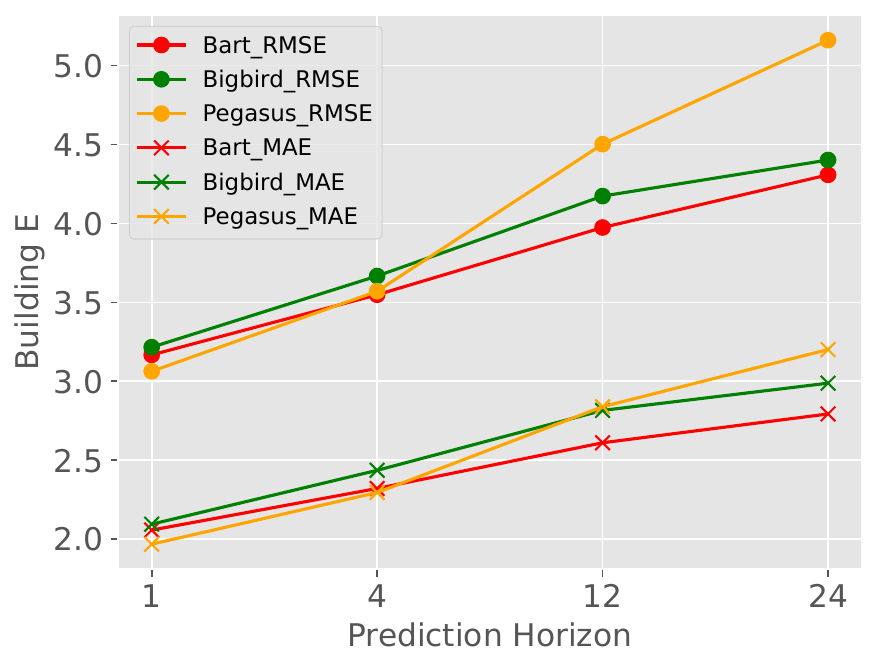}}
  \subfigure{\includegraphics[width=0.275\textwidth]{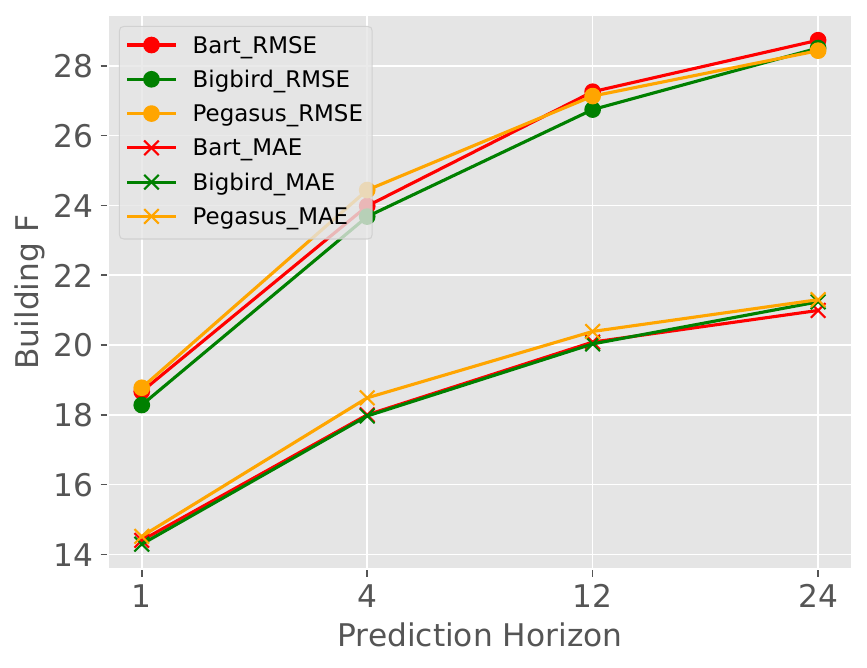}}
  
  \caption{The performance of language models with different prediction horizons.}
  \label{fig:subfigures}
\end{figure*}

\subsubsection{Zero-shot Performance}

In addition to the above evaluation of the language models, we also assess the performance through zero-shot setting. In our study, the zero-shot setting aims to evaluate the ability of the language model to generate reasonable predictions even without fine-tuning on the corresponding training data of one building.
Specifically, as listed in Table~\ref{tab:zero}, we fine-tuned the language models with the training set of one building. Instead of using the testing set of the same building, we directly use the fine-tuned model to generate predictions for the testings sets of other buildings.
In the table, we highlight the results that achieve similar or even better performance than the results reported in Table~\ref{tab:res}. As we can see, with our proposed pipeline, the language models can still yield plausible predictions for most of the buildings under the challenging zero-shot setting.
This evaluation provides insights into the language model's ability to generalize and transfer knowledge across different buildings. It also suggests that our proposed approach can be used to achieve reasonable forecasting results even for buildings without specific fine-tuning or buildings without enough recorded data to start with fine-tuning (\eg, cold start situation), which highlights the versatility and potential of language models in energy load forecasting.

\subsubsection{Different Prediction Horizons}

Furthermore, we investigate the effectiveness of our approach for different prediction lengths. We examine the performance of our method for different prediction lengths to assess its adaptability over varying time scales. Specifically, we set the prediction horizon $m=\{1, 4, 12, 24\}$ and the performance of three language models on 6 buildings are visualized in Figure~\ref{fig:subfigures}. For each building, the language model is only fine-tuned once (fine-tuned to generate the prediction of the next time step). The same fine-tuned model is used to yield forecasting for different horizons based on our autoregressive generation mechanism.
From the figure, although the error increases when the prediction horizon is enlarged (which is as expected), the results show that once the model is fine-tuned, it can be applied to arbitrary forecasting horizons.
The adaptability of language models in our forecasting pipeline across varying prediction horizons demonstrates the versatility and suitability of our method for real-world applications where dynamic, multi-step forecasting is essential.

\section{Conclusion}
We present a novel approach for energy load forecasting that leverages existing language models.
Through prompting, fine-tuning language models, and autoregressive prediction mechanism, our method enables accurate and dynamic predictions of energy consumption.
We have demonstrated the potential and the good performance of our proposed approach by evaluation with real-world data and comparisons against traditional numerical forecasting methods.
The zero-shot evaluation also reveals the ability of language models to generate reasonable predictions even without specific fine-tuning on a particular building. 
By harnessing the power of language models, our method provides a promising direction to unlock valuable insights for energy forecasting. 
Future research can focus on exploring prompt optimization to further improve the accuracy and applicability of language models in load forecasting.

\begin{acks}
We would like to acknowledge the support of Cisco’s National Industry Innovation Network (NIIN) Research Chair Program. We also highly appreciate the support from \href{https://c4net.com.au/}{C4NET} and \href{www.csiro.au}{CSIRO}.
\end{acks}

\bibliographystyle{ACM-Reference-Format}
\bibliography{sample-base}


\begin{thebibliography}{10}


\ifx \showCODEN    \undefined \def \showCODEN     #1{\unskip}     \fi
\ifx \showDOI      \undefined \def \showDOI       #1{#1}\fi
\ifx \showISBNx    \undefined \def \showISBNx     #1{\unskip}     \fi
\ifx \showISBNxiii \undefined \def \showISBNxiii  #1{\unskip}     \fi
\ifx \showISSN     \undefined \def \showISSN      #1{\unskip}     \fi
\ifx \showLCCN     \undefined \def \showLCCN      #1{\unskip}     \fi
\ifx \shownote     \undefined \def \shownote      #1{#1}          \fi
\ifx \showarticletitle \undefined \def \showarticletitle #1{#1}   \fi
\ifx \showURL      \undefined \def \showURL       {\relax}        \fi
\providecommand\bibfield[2]{#2}
\providecommand\bibinfo[2]{#2}
\providecommand\natexlab[1]{#1}
\providecommand\showeprint[2][]{arXiv:#2}

\bibitem[Lewis et~al\mbox{.}(2020)]%
        {bart}
\bibfield{author}{\bibinfo{person}{Mike Lewis}, \bibinfo{person}{Yinhan Liu},
  \bibinfo{person}{Naman Goyal}, \bibinfo{person}{Marjan Ghazvininejad},
  \bibinfo{person}{Abdelrahman Mohamed}, \bibinfo{person}{Omer Levy},
  \bibinfo{person}{Veselin Stoyanov}, {and} \bibinfo{person}{Luke
  Zettlemoyer}.} \bibinfo{year}{2020}\natexlab{}.
\newblock \showarticletitle{{BART:} Denoising Sequence-to-Sequence Pre-training
  for Natural Language Generation, Translation, and Comprehension}. In
  \bibinfo{booktitle}{\emph{Proceedings of the 58th Annual Meeting of the
  Association for Computational Linguistics, {ACL} 2020, Online, July 5-10,
  2020}}. \bibinfo{publisher}{Association for Computational Linguistics},
  \bibinfo{pages}{7871--7880}.
\newblock


\bibitem[Vaswani et~al\mbox{.}(2017)]%
        {vaswani2017attention}
\bibfield{author}{\bibinfo{person}{Ashish Vaswani}, \bibinfo{person}{Noam
  Shazeer}, \bibinfo{person}{Niki Parmar}, \bibinfo{person}{Jakob Uszkoreit},
  \bibinfo{person}{Llion Jones}, \bibinfo{person}{Aidan~N Gomez},
  \bibinfo{person}{{\L}ukasz Kaiser}, {and} \bibinfo{person}{Illia
  Polosukhin}.} \bibinfo{year}{2017}\natexlab{}.
\newblock \showarticletitle{Attention is all you need}.
\newblock \bibinfo{journal}{\emph{Advances in neural information processing
  systems}}  \bibinfo{volume}{30} (\bibinfo{year}{2017}).
\newblock


\bibitem[Xu et~al\mbox{.}(2021)]%
        {xu2021autoformer}
\bibfield{author}{\bibinfo{person}{Jiehui Xu}, \bibinfo{person}{Jianmin Wang},
  \bibinfo{person}{Mingsheng Long}, {et~al\mbox{.}}}
  \bibinfo{year}{2021}\natexlab{}.
\newblock \showarticletitle{Autoformer: Decomposition transformers with
  auto-correlation for long-term series forecasting}.
\newblock \bibinfo{journal}{\emph{Advances in Neural Information Processing
  Systems}}  \bibinfo{volume}{34} (\bibinfo{year}{2021}).
\newblock


\bibitem[Xue and Salim(2023)]%
        {xue2023promptcast}
\bibfield{author}{\bibinfo{person}{Hao Xue} {and} \bibinfo{person}{Flora~D.
  Salim}.} \bibinfo{year}{2023}\natexlab{}.
\newblock \bibinfo{title}{PromptCast: A New Prompt-based Learning Paradigm for
  Time Series Forecasting}.
\newblock
\newblock
\showeprint[arxiv]{2210.08964}


\bibitem[Xue et~al\mbox{.}(2022a)]%
        {xue2022translating}
\bibfield{author}{\bibinfo{person}{Hao Xue}, \bibinfo{person}{Flora~D Salim},
  \bibinfo{person}{Yongli Ren}, {and} \bibinfo{person}{Charles~LA Clarke}.}
  \bibinfo{year}{2022}\natexlab{a}.
\newblock \showarticletitle{Translating Human Mobility Forecasting through
  Natural Language Generation}. In \bibinfo{booktitle}{\emph{Proceedings of the
  Fifteenth ACM International Conference on Web Search and Data Mining}}.
  \bibinfo{pages}{1224--1233}.
\newblock


\bibitem[Xue et~al\mbox{.}(2022b)]%
        {xue2022leveraging}
\bibfield{author}{\bibinfo{person}{Hao Xue}, \bibinfo{person}{Bhanu~Prakash
  Voutharoja}, {and} \bibinfo{person}{Flora~D Salim}.}
  \bibinfo{year}{2022}\natexlab{b}.
\newblock \showarticletitle{Leveraging language foundation models for human
  mobility forecasting}. In \bibinfo{booktitle}{\emph{Proceedings of the 30th
  International Conference on Advances in Geographic Information Systems}}.
\newblock


\bibitem[Zaheer et~al\mbox{.}(2020)]%
        {zaheer2020big}
\bibfield{author}{\bibinfo{person}{Manzil Zaheer}, \bibinfo{person}{Guru
  Guruganesh}, \bibinfo{person}{Kumar~Avinava Dubey}, \bibinfo{person}{Joshua
  Ainslie}, \bibinfo{person}{Chris Alberti}, \bibinfo{person}{Santiago
  Ontanon}, \bibinfo{person}{Philip Pham}, \bibinfo{person}{Anirudh Ravula},
  \bibinfo{person}{Qifan Wang}, \bibinfo{person}{Li Yang}, {et~al\mbox{.}}}
  \bibinfo{year}{2020}\natexlab{}.
\newblock \showarticletitle{Big bird: Transformers for longer sequences}.
\newblock \bibinfo{journal}{\emph{Advances in Neural Information Processing
  Systems}}  \bibinfo{volume}{33} (\bibinfo{year}{2020}),
  \bibinfo{pages}{17283--17297}.
\newblock


\bibitem[Zhang et~al\mbox{.}(2020)]%
        {zhang2020pegasus}
\bibfield{author}{\bibinfo{person}{Jingqing Zhang}, \bibinfo{person}{Yao Zhao},
  \bibinfo{person}{Mohammad Saleh}, {and} \bibinfo{person}{Peter Liu}.}
  \bibinfo{year}{2020}\natexlab{}.
\newblock \showarticletitle{Pegasus: Pre-training with extracted gap-sentences
  for abstractive summarization}. In \bibinfo{booktitle}{\emph{International
  Conference on Machine Learning}}. PMLR, \bibinfo{pages}{11328--11339}.
\newblock


\bibitem[Zhou et~al\mbox{.}(2021)]%
        {zhou2021informer}
\bibfield{author}{\bibinfo{person}{Haoyi Zhou}, \bibinfo{person}{Shanghang
  Zhang}, \bibinfo{person}{Jieqi Peng}, \bibinfo{person}{Shuai Zhang},
  \bibinfo{person}{Jianxin Li}, \bibinfo{person}{Hui Xiong}, {and}
  \bibinfo{person}{Wancai Zhang}.} \bibinfo{year}{2021}\natexlab{}.
\newblock \showarticletitle{Informer: Beyond efficient transformer for long
  sequence time-series forecasting}. In \bibinfo{booktitle}{\emph{Proceedings
  of AAAI}}.
\newblock


\bibitem[Zhou et~al\mbox{.}(2022)]%
        {FEDformer}
\bibfield{author}{\bibinfo{person}{Tian Zhou}, \bibinfo{person}{Ziqing Ma},
  \bibinfo{person}{Qingsong Wen}, \bibinfo{person}{Xue Wang},
  \bibinfo{person}{Liang Sun}, {and} \bibinfo{person}{Rong Jin}.}
  \bibinfo{year}{2022}\natexlab{}.
\newblock \showarticletitle{FEDformer: Frequency Enhanced Decomposed
  Transformer for Long-term Series Forecasting}. In
  \bibinfo{booktitle}{\emph{International Conference on Machine Learning,
  {ICML} 2022, 17-23 July 2022, Baltimore, Maryland, {USA}}}
  \emph{(\bibinfo{series}{Proceedings of Machine Learning Research},
  Vol.~\bibinfo{volume}{162})}. \bibinfo{publisher}{{PMLR}},
  \bibinfo{pages}{27268--27286}.
\newblock


\end{thebibliography}


\end{document}